# The Problem of Adhesion Methods and Locomotion Mechanism Development for Wall-Climbing Robots

N. S. Vlasova and N. V. Bykov
Bauman Moscow State Technical University, Moscow, Russia
Email: vlasovans@bmstu.ru, bykov@bmstu.ru

*Abstract* – This review considers a problem in the development of mobile robot adhesion methods with vertical surfaces and the appropriate locomotion mechanism design. The evolution of adhesion methods for wall-climbing robots (based on friction, magnetic forces, air pressure, electrostatic adhesion, molecular forces, rheological properties of fluids and their combinations) and their locomotion principles (wheeled, tracked, walking, sliding framed and hybrid) is studied. Wall-climbing robots are classified according to the applications, adhesion methods and locomotion mechanisms. The advantages and disadvantages of various adhesion methods and locomotion mechanisms are analyzed in terms of mobility, noiselessness, autonomy and energy efficiency. Focus is placed on the physical and technical aspects of the adhesion methods and the possibility of combining adhesion and locomotion methods.

## I. INTRODUCTION

The development of wall-climbing robots (WCR) is one of the challenging issues of modern robotics. The first WCRs appeared in the mid-1960s [1]; in the 1990s, research and development on WCR methods and mechanisms began to extensively progress, and by 2005, there were more than 200 experimental models and prototypes [2]. In the 2000s, with magnetic and vacuum adhesion, new methods of adhesion were developed, which led to an increase in scientific research and publications in this direction [3-16]. However, at present, there is no universal robot that satisfies the operational conditions in various environments and on all surfaces. The main reason is the features of adhesion methods and locomotion mechanisms for climbing, which are analyzed in this paper.

The general idea of the robot assignment is to solve tasks that are expensive, potentially dangerous, or difficult to accomplish, including for environmental reasons. WCRs created to address problems of nondestructive inspection of construction, industrial and technical facilities [8, 10, 13, 16-37] are the most widespread. The second group of tasks solved using WCRs includes cleaning or painting objects that are difficult to access or potentially dangerous to humans [5, 6, 8, 10, 15, 18, 34, 38-46]. The third group of tasks for WCRs is in the field of security, where the WCRs provide assistance to military police and special services [5, 8, 10, 25, 34, 37, 42, 47-50].

## II. ADHESION METHODS

The adhesion methods determine the WCR's ability to effectively attach, hold and detach from a surface. The currently existing adhesion methods can be divided into groups according to the nature of the adhesion force (Fig. 1). The adhesion methods can also be divided into active and passive depending on whether energy is expended in creating a holding force.

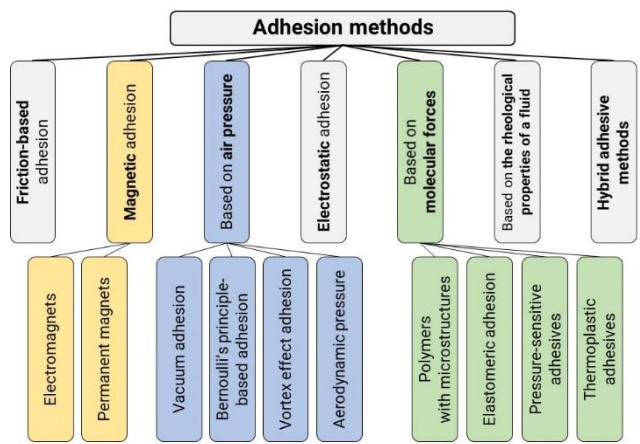

Figure 1. Classification of adhesion methods according to the nature of the adhesion force

### A. Friction-based adhesion

This type of adhesion is realized in WCRs using mechanical elements (claws, hooks, spikes) that are embedded in deformable surfaces or cling to the roughnesses on uneven surfaces [3, 4, 10, 12, 26, 27, 35, 48, 49, 51-54]. The mechanical element scale varies in the range of 1-100 μm depending on the operating surface type and quality. The nature of the WCR interaction with the operating surface, which accounts for the locomotion mechanism, determines the number of mechanical elements and their design, material and shape, which are best suited for moving along vertical surfaces.

### B. Magnetic adhesion

This adhesion method is used when a WCR moves along ferromagnetic surfaces or structures with large ferromagnetic elements (such as reinforced concrete structures). It provides a high robot maneuverability and larger holding force than other adhesion methods. Magnetic adhesion can be provided in an active or passive manner. In the first case, electromagnets are used [3, 4, 6, 8-10, 28, 55-57]; in the second case, permanent magnets are used [3, 4, 6, 8-10, 14, 20, 22, 24, 29-31, 36, 58, 59].

### C. Adhesion methods based on air pressure

Adhesion methods based on air pressure include vacuum adhesion, Bernoulli's principle-based adhesion, vortex effect (Ranque-Hilsch effect) adhesion and aerodynamic pressure. These methods are only realizable in the atmosphere and are nonusable in space.

Vacuum adhesion mechanisms are some of the most usable mechanisms in WCRs. The holding force in this case can be passively or actively formed.

Passive vacuum adhesion does not require energy for the holding force formation and is realized with suction cups [3, 4, 6, 8, 10, 16, 60-65]. In this case, it is necessary to provide a preload to press the suction cups to the surface and an additional force for their separation. Both mechanisms should be provided by the design of a locomotion mechanism [62].

Active vacuum adhesion [3, 4, 6, 8, 10, 13, 16, 18, 23, 37-40, 42, 66-69] is provided by external or internal (on-board) vacuum generators. In [42], a WCR prototype attached with four suction cups to a vertical surface is considered. The vacuum in the cups is formed with a liquid vacuum pump that operates on the Venturi effect. The operating medium (water, air), whose kinetic energy is used to provide a holding force, can also be directly used to solve tasks by robots such as window cleaning.

Bernoulli's principle is widely used in industrial robots for the contactless gripper operation, but using this principle to hold the robot on a vertical surface is a slightly new phenomenon. The WCR blows air to the wall surface; as a result, the greater external pressure holds it on the wall [70]. The experimental model in [70] can move in any direction and overcome small obstacles such as joints between tiles on a wall or cracks in a tree. Its payload capacity is 0.5 kg with an own weight of 0.234 kg.

The robot can be held on a vertical surface with the vortex effect (Ranque-Hilsch effect) [4, 16, 19, 32, 47, 71, 72]. Similar to the previous mechanism, this principle is used in contactless grippers in industrial robotics. One method to provide a vortex effect is mechanical: an impeller connected to the drive is installed in the closed cavity of the vacuum (suction) chamber. When the impeller is retwisted in the central part of the suction chamber, an area with reduced pressure is formed and makes the robot hold on the surface. Low-power-consumption engines are sufficient to rotate the impeller for a given WCR energy efficiency. The negative pressure inside the suction chamber can be controlled by changing the rotation speed of the impeller. The payload capacity of WCRs that operate on this effect is relatively high: for example, in the prototype [72], it is 0.5 kg, and the weight of the robot is 0.12 kg.

WCRs that operate based on aerodynamic pressure are held on a vertical surface by the pressure force formed by rotating propellers [21, 33, 43, 73]. In these WCRs with an active locomotion mechanism, the propellers act as an adhesion mechanism and a source of driving force to move the robot along the surface.

### D. Electrostatic adhesion method

This adhesion method is based on an electrostatic force created by high voltage (0.5-5 kV), which is applied to a flexible electrode panel [4, 12, 34, 74-79]. The panel consists of at least two sets of independent electrodes with different potentials; the electrodes are deposited on one polymer film and coated with another polymer film to provide electrical insulation. When alternating positive and negative charges are induced at the adjacent electrodes, the electric fields create opposite charges on the wall surface. Thus, they cause electrostatic adhesion between the electrode panel and the wall surface. The strength of adhesion directly depends on the electrode panel area, its materials and the magnitude of the applied voltage [78].

WCRs that operate on the electrostatic adhesion have low weight and small dimensions due to the simple adhesion mechanism construction and require low power consumption in the resting state. For example, an electrode panel with a 40-cm$^2$ area requires 0.25-mW of power at currents of 10-20 nA per N of weight [79]. The dynamic properties of the WCRs with electrostatic adhesion are currently unsatisfactory. First, the turning movement is difficult for such robots, since it causes a dynamic change of the electric field and a decrease in the holding electrostatic force [75]. Second, the maximum movement speed is limited to the effect of charge relaxation [76].

### E. Adhesion based on molecular forces

These adhesion methods are most widely realized in "dry" adhesion, but there is also a "wet" adhesion.

Dry adhesion is realized by having materials interacting at the molecular level with the operating surface, such as synthetic elastomers or materials with micro- or nanostructure arrays.

The adhesion of polymers with microstructures simulates the morphology of a gecko's paws, which have a huge number of micro- and nanohairs that adhere to the supporting surface due to van der Waals forces [80, 81]. This principle of a holding force formation is realized in WCR using materials that contain arrays of microstructures [4, 8, 10, 12, 82-89]. The shape, size, material, and location of microfibers in these arrays determine the adhesion rate [88, 89]. In addition, WCRs using polymers with microfibers are being actively investigated for space applications [82, 84].

Elastomeric adhesion (synthetic "dry" adhesion) [4, 8, 10, 12, 14, 25, 41, 81, 89, 90] can be a substitute for complex and expensive polymers with microstructures. In this case, simpler and cheaper adhesive materials with small elasticity moduli are used, and the adhesive force is proportional to the area of the elastomer contact surface.

Adhesion in the wet state has been performed in experimental WCR models [91-93]. These WCRs use adhesives in the liquid state to adhere to the operating surface. These materials serve as an interlayer between the surface and the contact pad of the robot. Pressure-sensitive adhesives (passive adhesion) [91] and thermoplastic adhesives (active adhesion) can be used. In the second case, the robot requires a heating element to adhere to the surface. It can be a built-in automatic hot melt device that dispenses glue to the surface [92] or a heating element embedded in a robot contact pad made of thermoplastic glue with a cooling element [93].

For mini robots (characteristic size < 100 mm), a "reverse" version of adhesion in the wet state with the use of glue has been proposed: the adhesive is applied in advance on the operating surface, upon which the robot then climbs [94]. Studies have shown that with similar scales of robots, this approach provides efficient movement on vertical surfaces.

*F. Adhesion based on the rheological properties of a fluid*

Although these methods are used to form a holding force, they have not found a wide range of application. For example, the snail robot with an original locomotion mechanism [95] uses a non-Newtonian fluid, whose characteristics are similar to snail slime, to move along walls and ceilings. Another example is the magnetorheological fluid application for WCR adhesion with the surface [96], which enables the robot to move along surfaces of different types.

*G. Hybrid adhesion methods*

These methods are a combination of two or more described methods [50, 97, 98]. Hybrid adhesion methods can improve the adhesion efficiency and adaptability of WCRs to various types of surfaces.

III. LOCOMOTION MECHANISMS

The locomotion mechanisms of WCRs are the technical methods of implementing their movement. These mechanisms can be divided into passive and active. Passive mechanisms use a device that does not belong to the WCR, such as a cable, rope or rail device [3, 6, 12, 15, 39, 43-46]. The movement of such robots is provided by roof cables, winches and other devices that provide their movement. Such mechanisms are commonly used to move large and heavy robots that can carry a large payload. However, the WCRs that move in such manners are not autonomous and maneuverable and have a relatively low speed of movement, so they are not considered later in the article.

The active locomotion mechanisms of WCRs can be divided into six groups: wheeled, tracked, walking, sliding framed, hybrid and specific, which cannot be attributed to any of the previous groups. For example, the last group includes robots that imitate the movement of a snail [95] or a caterpillar [65, 69, 79] or a jumping robot [54].

The wheeled locomotion mechanism is one of the most common for WCRs [3-5, 8-10, 12, 13, 16, 19-21, 24, 30, 51, 57, 58, 70, 71, 91]. Each link of a wheeled robot has several wheels driven by engines. The maneuverability of wheeled robots can be provided by differential drives or omnidirectional wheels. Based on the number of wheels, these locomotion mechanisms can be divided into two-, three-, four-, six- and multi-wheeled (with more than eight wheels) drives. Wheeled WCRs are used on surfaces with medium roughness to avoid slipping. However, there is another solution: to select the wheel material depending on the type of surface. In [70], the robot easily (using a single screw) mounts the necessary wheels for the conditions of a specific task. Compared to other locomotion mechanisms, wheeled WCRs have good mobility, a simple drive mechanism (usually electric) and the ability to achieve high speeds with low power consumption. However, they have limitations when they must overcome obstacles with a sharp height drop, which exceeds one third of the wheel diameter.

The tracked locomotion mechanism is also widely used in WCRs [3, 4, 6, 8, 9, 12, 14, 22, 29, 36, 61, 62, 65-67, 72, 75-77, 79, 83, 85, 97]. These robots move by tracks that are driven by drive mechanisms. Tracked WCRs have good and continuous contact with the operating surface. This mechanism provides them with good friction and driving force to resist slip and enables them to move along uneven and soft surfaces and overcome obstacles; they have an advantage in speed of movement compared to the walking and sliding framed mechanisms and can accommodate large payloads. However, they consume more energy, reach lower speeds and are less maneuverable than the wheeled robots.

The walking locomotion mechanism [3-5, 8, 10, 12, 16, 18, 28, 34, 35, 38, 48, 49, 52, 53, 55, 63, 64, 68, 86, 92, 93, 96] has analogs in nature by design and kinematics and simulates the gait of living beings. Based on the number of walking links ("legs"), such WCRs can be divided into two-, four-, five-, six- and multilegged types. The high mobility of the walking links is the advantage of walking robots, which enables them to move on uneven surfaces with high obstacles and in unstructured environments. However, due to the intermittent movement, such robots have relatively low movement speeds, consume large amounts of energy, are difficult to control in terms of gait, and do not have good reliability and stability when moving along a vertical surface.

The sliding framed locomotion mechanisms [3-5, 9, 10, 12, 14, 16, 23, 27, 39-42, 74] consist of two moving parts, which translationally or rotationally move relative to each other. This mechanism is simple in operation and control compared to the locomotion mechanisms of other WCRs and can provide a large step size. However, despite its simplicity, its disadvantages are the large size of the body (frame), which does not enable movement in confined spaces, low speed, low ability to overcome obstacles and difficulties in maneuvering.

Hybrid locomotion mechanisms are combinations of several of the above mechanisms [4, 12, 37, 50, 56]. Such combinations combine the features of the mechanisms to increase the speed of movement and the possibility of overcoming obstacles, but they complicate the WCR design and control.

## IV. COMBINATIONS OF LOCOMOTION MECHANISMS AND ADHESION METHODS

In many cases, the locomotion mechanism of the WCR is combined with the adhesion method. For example, suction pads [60-62, 66, 67] or permanent magnets [14, 29] can be placed on the robot track; electrodes for electrostatic adhesion formation can be the track itself [74-77, 97]. In wheeled WCRs, the wheels can be magnetic [20, 56, 58, 59], have mechanical elements (spikes) [51], or consist of adhesive "feet" [25, 82]. Such combinations of two mechanisms simplify the design of the mobile robots, reduce the size and weight, improve the control quality and reduce the power consumption.

Table 1 summarizes the applicability and compatibility of adhesion methods with locomotion mechanisms, which are collected from the existing prototypes and experimental samples of WCRs.

Table 2 shows the generalized values of the main parameters of WCRs for various adhesion methods.

Table 3 shows the comparative quality indicators for WCRs with different adhesion methods.

TABLE I. APPLICABILITY AND COMPATIBILITY OF ADHESION METHODS AND LOCOMOTION MECHANISMS OF EXISTING PROTOTYPES AND EXPERIMENTAL SAMPLES OF WCRS

| Adhesion methods and locomotion mechanisms | | Tracked | Wheeled | Walking | Sliding framed | Hybrid | Specific |
|---|---|---|---|---|---|---|---|
| | Friction based | - | + | + | - | - | + |
| Magnetic | Electromagnets | - | + | + | + | + | - |
| | Permanent magnets | + | + | + | - | - | - |
| Air Pressure | Active vacuum | + | + | + | + | + | - |
| | Passive vacuum | + | - | + | - | - | + |
| | Vortex effect | + | + | - | - | + | - |
| | Bernoulli's principle | - | + | - | - | - | - |
| | Aerodynamic pressure | - | + | - | - | - | - |
| | Electrostatic adhesion | + | - | + | + | - | + |
| Dry adhesion | Polymers with microstructures | + | - | + | - | - | - |
| | Elastomeric | + | + | + | + | - | - |
| Wet adhesion | Glues | - | + | + | - | - | - |
| | Thermoplastics | - | - | + | - | - | - |
| | Based on rheological properties of liquids | - | - | + | - | - | + |

TABLE II. MAIN PARAMETERS OF WCRS WITH DIFFERENT ADHESION METHODS

| Adhesion methods and WCR parameters | | Payload to WCR mass ratio | WCR mass (kg) | Typical length scales (cm) | Velocity (cm/s) |
|---|---|---|---|---|---|
| | Friction based | 0.5…1 | 0.4…5.5 | 10…100 | 2…20 |
| Magnetic | Electromagnets | 1 | 0.6…100 | 20…110 | 14 |
| | Permanent magnets | 0.9…3.8 | 1…10 | 30 | 8 |
| Air pressure | Active vacuum | 0.2…2 | 1…70 | 20…150 | 0.4…25 |
| | Passive vacuum | – | 0.03…1.2 | 12…50 | 1…90 |
| | Vortex effect | 0.3…4.2 | 0.15…5 | 15…50 | 10…20 |
| | Bernoulli's principle | 2.1 | 0.2 | 22 | – |
| | Aerodynamic pressure | 1.6 | 0.6 | 35…210 | 20 |
| | Electrostatic adhesion | 2…3 | 0.1…0.7 | 10…60 | 0.2…15 |
| Dry adhesion | Polymers with microstructures | 0…1.5 | 0.08…0.6 | 20…60 | 0.1…40 |
| | Elastomeric | 0…2.6 | 0.06…0.15 | 5…20 | 0.8…12 |
| Wet adhesion | Glues | – | 0.08 | 10 | 0.4…5 |
| | Thermoplastics | 0…7 | 0.6…1.4 | 15…30 | 0.04…0.8 |

TABLE III. QUALITATIVE INDICATORS FOR WCRs WITH DIFFERENT ADHESION METHODS

| Adhesion methods and WCR parameters | | Ability to operate on | | | Repeatability of operation | Noise level |
|---|---|---|---|---|---|---|
| | | various materials | dirty surfaces | uneven surfaces | | |
| | Friction based | □ | ● | ● | * | * |
| Magnetic | Electromagnets | * | □ | □ | ● | * |
| | Permanent magnets | * | □ | □ | ● | * |
| Air pressure | Active vacuum | ● | □ | □ | ● | ● |
| | Passive vacuum | ● | □ | □ | ● | * |
| | Vortex effect | ● | □ | ● | ● | ● |
| | Bernoulli's principle | ● | ● | ● | ● | ● |
| | Aerodynamic pressure | ● | ● | ● | ● | ● |
| | Electrostatic adhesion | ● | ● | ● | ● | * |
| Dry adhesion | Polymers with microstructures | ● | * | □ | * | * |
| | Elastomeric | ● | * | □ | * | * |
| Wet adhesion | Glues | ● | * | ● | * | * |
| | Thermoplastics | ● | * | ● | * | * |

● – high, □ – medium, * – low

Most WCRs constructively consist of one link (body) and include an adhesion mechanism and a locomotion mechanism, which realizes the necessary and sufficient movement of the robot to solve a specific task.

However, the WCRs must have great mobility, maneuverability, and ability to overcome the transitions between vertical and horizontal surfaces through internal or external angles of different sizes, overcome obstacles and move on arbitrarily oriented surfaces when the WCRs are designed to move on complex surfaces, which are arbitrarily oriented in space. For this purpose, two- [14, 17, 36, 71, 79, 85], three- [14, 58, 66] and multilink [5, 14] WCRs are used.

## V. CONCLUSION

The analysis of the methods and mechanisms of robot movements demonstrates a great variety in applied adhesion methods, which enable robots to be held on vertical surfaces, and locomotion mechanisms, which provide their movement along operating surfaces. Each of the considered methods and mechanisms has advantages and limitations in application, which makes it impossible to create a universal wall-climbing platform that is equally well suited for solving various tasks in different conditions. The tendency of bionic and hybrid method development has emerged in recent years in the WCR design process.

At the current stage of development of WCRs, the following problems can be identified, the solution of which will enable significant advances. Most adhesion methods do not enable the implementation of an autonomous robot, even fewer methods enable the possibility of passive adhesion, which does not require a specially organized external energy supply. In the scope of a single adhesion method, it appears impossible to develop a robot that can move across a wide range of surface types. Based on these features, hybrid adhesion methods should be developed to address a number of practical problems.

ACKNOWLEDGMENT

The research was supported by the Russian Foundation for Basic Research (RFBR), grant No. 16-29-09596 ofi-m.